Scaling up Greedy Causal Search for Continuous Variables[1]

Joseph D. Ramsey

Technical Report
Center for Causal Discovery
Pittsburgh, PA

11/11/2015

Abstract

**As standardly implemented in R or the Tetrad program, causal search algorithms used most widely or effectively by scientists have severe dimensionality constraints that make them inappropriate for big data problems without sacrificing accuracy. However, implementation improvements are possible. We explore optimizations for the Greedy Equivalence Search that allow search on 50,000-variable problems in 13 minutes for sparse models with 1000 samples on a four-processor, 16G laptop computer. We finish a problem with 1000 samples on 1,000,000 variables in 18 hours for sparse models on a supercomputer node at the Pittsburgh Supercomputing Center with 40 processors and 384 G RAM. The same algorithm can be applied to discrete data, with a slower discrete score, though the discrete implementation currently does not scale as well in our experiments; we have managed to scale up to about 10,000 variables in sparse models with 1000 samples.**

1. Introduction.

Under the auspices of the Center for Causal Discovery, we recently scaled the Greedy Equivalence Search (GES, Meek 1995; Chickering 2003) algorithm to handle variable sets in the tens of thousands on a laptop computer. This is for sparse models with the number of edges equal to the number of nodes and all continuous variables. On a supercomputer node at the Pittsburgh Supercomputing Center, it will handle variable sets in the hundreds of thousands of variables. The largest sparse model we have tested it on used 1,000,000 variables, which is an exponentially significant improvement over our previous work in which we have could only been able to analyze models of 5,000 variables, requiring days of computation. It seems fitting, therefore, to stop for a moment. Here we document how this speedup was achieved and to make the code available to the community.

The revised GES algorithm we FGS for "Fast Greedy Search" We discuss FGS in stages. First, we review the basic assumptions and structure of the GES algorithm as given in Chickering

[1] Research supported by the National Institutes of Health under Award Number U54HG008540. The content is solely the responsibility of the author and does not necessarily represent the official views of the National Institutes of Health. Special thanks to Clark Glymour, Peter Spirtes, and Erich Kummerfeld for algorithmic help and to Nick Nystrom and John Urbanic for help in running the algorithm on Pittsburgh Computing Center machines.

(2002). Next, we go over the strategic optimizations used for the FGS implementation. We will take for granted that coders will have used basic code profiling and established programming techniques for optimization and therefore will not discuss these, even though they contribute significantly to scale-up. Next, we will define a simulation framework and give results for both a laptop computer and a supercomputer node. We end with a brief discussion.

2. Assumptions.

GES assumes the data are generated from parameterizations of a true causal model characterized by a directed acyclic graph (DAG), in which directed edges X→Y indicate that X is a cause of Y. A directed graph over a set of variables V consists entirely of directed edges. A path is a concatenation of edges, e.g. X→Y←Z; a directed path is a path consisting entirely of directed edges pointing in the same direction. A directed graph is acyclic if it contains no path from a variable to itself. Some variables may be measured; those that are unmeasured are called latent variables. GES additionally assumes that the graph over the measured variables does not contain variables with a common latent cause: that is, unmeasured variable L such that X<-L->Y for X and Y measured. Under these assumptions, GES (like several other algorithms) produces a pattern or CPDAG, which is a mixed graph consisting of both directed and undirected edges representing an equivalence class of DAGs.

An unshielded triple is a path X*-*Y*-*Z of variables (where '*' is either a tail endpoint or an arrow endpoint), where X and Z are not adjacent, such as a path of the form X->Y<-Z. The pattern will have oriented each unshielded triple in the graph, together with as many other orientations that one can make without introducing new unshielded triples. All other edges will remain unoriented. A process for obtaining DAGs in the pattern suggests itself: one simply takes an edge that is unoriented in the pattern, gives it a arbitrary direction, and then orients as many other edges as possible without creating new unshielded colliders. One repeats this process until the resulting graph is a DAG.

Importantly, all DAGs in the pattern are statistically indistinguishable, in the sense that there's no reason to prefer any of them over another based on facts about conditional independence. If variables relate to one another by linear functions using Gaussian errors or through discrete multinomial conditional probability tables, these DAG are indistinguishable, but they also cannot be distinguished if conditional independence is calculated (even correctly) for arbitrary functional forms. (Other algorithms using as information more than just conditional independence may be able to distinguish suitability of DAGs inside of patterns.) GES for continuous variables has been worked out primarily for the linear, Gaussian case, so we will examine that below. It is interesting that GES does well at inferring causal inference even when disturbances are moderately non-Gaussian or connection functions moderately nonlinear; one may enhance this effect success by using the Liu et al. nonparanormal transform (Liu et al., 2009).

3. Performance Measures.

Algorithm performance is reported primarily as elapsed time and accuracy. In all cases, the algorithm easily fits inside available memory, though the process uses virtual memory sometimes on a desktop computer There will be four dimensions to the measurement of accuracy. We calculate for adjacencies and arrow orientations the number of true positives (TP), the number of false positives (FP), and the number of false negatives (FN). Then for adjacencies and arrow orientations respectively, we calculate the precision (TP / (TP + FP)) and recall (TP / (TP + FN)). An ideal algorithm will have precision and recall close to 100%.

To be clear about false positive and negative arrows, consider some simple examples. Let the true model be X->Y; the pattern for this is X--Y. If the output model is X<-Y or X->Y, in each case, this will count as an arrow error. Likewise, if there is, in fact, no edge between X and Y yet the output model is X->Y or X<-Y, this will count as both an adjacency error and a direction error. And so on for other cases.

It is worth noting up front that in the tested configuration, increasing the amount of RAM, even significantly, does not necessarily improve running time or the ability of the process to finish without throwing exceptions. The reason is that covariances are calculated on the fly, so that one only needs to store the data set in memory, together with requisite fields in the heap for processing the algorithm. Calculating covariances on the fly is a deliberate choice, since precalculating the covariance matrix severely limits the number of variables that can be analyzed. Covariance matrices cannot be stored in memory for either the MacBook Pro or the supercomputer nodes for problems of even moderate size; they are therefore not stored in matrix form but are rather computed on the fly as needed.

4. Overview of GES Algorithm.

GES uses scoring to produce an output pattern, adding or removing one edge at a time in the space of patterns until it can make no more such changes. The algorithm breaks down into two pieces. In the first piece, the maximum scoring edge among all possible edges satisfying certain conditions is added to the pattern, and the orientation of the pattern is adjusted for the addition. In the second piece, the maximum scoring edge among all possible edges satisfying certain conditions is removed, and the orientation of the pattern is adjusted for the removal. In either case, the score of the model is monotonically increased.

The score used in the algorithm must be decomposable (Chickering 2002): that is, the score of the model must be the sum of scores for each variable given its parents in the graph. The log of the Bayesian decomposition of the model is a decomposition of this nature. Let P(M) = P(X1 | Pa(X1)) P(X2 | Pa(X2) ... P(Xn | Pa(Xn)), where X1,..., Xn are variables in graph G. Let Pa(X) be the parents of X in graph G, and let P(M) be the probability of the model. Then taking logs, we have ln P(M) = ln P(X1 | Pa(X1)) + ... + ln P(Xn | Pa(Xn)). We assume that the probability of the model is decomposable in this way. This formulation translates directly into a formulation involving likelihoods: L = P(M | D) = ln P(X1 | Pa(X1), D) + ... + ln P(Xn | Pa(Xn), D), where D is a data set for variables X1,.., Xn. Since the likelihood of a variable given its parents in the linear, Gaussian case is equal to -n ln s + C for sample size n, residual variance after regressing X onto its parents s, and constant C, we have that the BIC score is equal to:

BIC = 2 L - k ln n = -n ln s - k ln n + C

(Schwarz, 1978). Here, k is the number of degrees of freedom of the model. Notice that BIC is decomposable since L is decomposable, and the penalty k ln n is the sum of penalties for individual BIC scores of variables given their parents. This score is adjusted in the direction of stringency for large models as follows:

BIC2 = 2 L - c k ln n

where k = 2p + 1 for a variable given its parents Y1,..., Yp, and c is a penalty discount chosen empirically to be by default 2.

As indicated earlier, the forward phase of the algorithm consists of a loop for adding edges to the model, and the backward phase consists of a loop for removing edges from the model. The forward phase begins with the problem of determining which single edge to add to the model, to get things started. Deciding which initial edge to add is a very expensive step, especially for large models, and is quadratic in the number of variables. For each pair of variables <X, Y>, we must score the model consisting of Y with no parents, yielding score s1, and then the model consisting of Y with X as a parent, yielding score s2. The difference between these scores, s1 - s2, may be positive or not. If it is not positive, the edge X->Y is not added to the model. If it is positive, it is considered as a candidate to add to the model. Out of all such candidates to be added to the model, we choose the one with the highest difference in score and add this edge to the model. Note that the difference in score for X->Y is the same as the difference in score for Y->X; one only needs to test one of these. After this step, the graph is reverted to its pattern, in this case, by rendering the arrow X->Y as undirected X--Y.

For the next phase, for each edge X->Y considered for addition, one needs two sets for each edge: NaYX and T (Chickering 2002). NaYX is the set of all Z such that Z--Y and such that Z is adjacent to X. T is the set of all Z such that Z--Y and such that Z is not adjacent to X. Together, NaYX and T comprise all of the variables such that X--T. {Y} U S U NaYX must not be a clique, and adding X->Y to the graph must not result in a cycle. That is, there must be no semi-directed path from Y to X, where a semi-directed path is a path from X to Y that consists entirely of undirected edges and edges pointing toward Y. If these checks pass, then for each subset S of T, we score Y with all of S as parents, yielding score s1, and Y with all of S as parents and also with X as a parent, yielding score s2; the score difference is s1 - s2. If this difference in score is positive and is the highest out of all such calculated scores, the edge X->Y (with subset S) is added to the graph. Once this finishes, one reorients S'->Y for each S' in S. Note that this reorientation of S' into S is how colliders get oriented in the graph. The graph is reverted to its pattern by rendering as undirected all edges not involved in unshielded colliders and then applying the Meek rules. Note that this step might orient edges in the graph that were previously unoriented and might unorient some edges that the algorithm did not orient previously.

The backward step is similar to the forward step, with some modifications. First, subsets H of NaYX are calculated instead of subsets of T. Second, the clique check is for NaYX \ H. Third, the cycle check is not necessary. And fourth, when the algorithm removes X->Y, it renders edges Y--H' as Y->H' for each H' in H.

5. Optimizations for FGS.

Profiling a naïve implementation of the GES algorithm reveals two basic bottlenecks. First, there is considerable redundancy in scoring. The initial step of adding a single edge to the graph is quadratic in the number of variables, and adding each additional edge to the graph is quadratic in the number of variables. Most of the scoring in each of these steps is repeated many times over. However, much of the redundancy in this scoring can be reduced. Second, even if one reduces scoring to a minimum, the process of calculating BIC scores itself is extraordinarily expensive. While we may optimize this step by calculating regressions directly from the covariance matrix, it is still expensive.

There are many ways to address the redundant scoring problem; our method is as follows. We keep a list L of score differences, along with associated information. The associated information consists of a list of structures of "arrows": A = <d, X->Y, S, NaYX> for the forward step. Here, d is the score difference, X->Y is the edge that one could add to the graph, and S is a subset of the set of variables adjacent to Y by undirected edges and adjacent also to X. NaYX is the set of variables adjacent to Y by undirected edges but not adjacent to X. The list L need only contain arrows with positive score differences; we exclude arrows with non-positive score differences from L. L is kept in sorted order from high to low by differences. (There are collection classes in Java that keep data in sorted order; a collection class available in Java for this called SortedSet.) It is important to note that the edge X->Y may appear in more than one arrow since we calculate such arrows for each subset of the T set described above. However, when we remove an edge from the list, all other edges with different subsets of T must be removed as well to maintain the consistent property that the list of edges consists of a subset of edges that one would have calculated at a particular stage of the algorithm. We always add edge with the highest score to the graph, so long as it is a legal edge (i.e., passes the clique and cyclicity tests). When we add an edge to the graph, say X->Y, possibly reverted to X--Y, scores for edges into X or Y, say, W->X, must be re-scored. The reason is that they may have a new parent, say X->Y<-W, resulting in two parents, where before there was just the one parent X of Y. The full rescoring step takes every node W in the graph other than X or Y and re-scores the edge W->X, and if X is adjacent to Y, X->W, and similarly for W->Y and Y->W. This rescoring is an expensive step, but it is linear in the number of variables rather than quadratic, so there is significant time saving.

The revised algorithm, therefore, starts a list sorted by score difference of arrows. After an extensive search, the first edge X->Y is added to the graph, and the graph reverted to its pattern. Then, in the manner described above, new edges are added to L for variables adjacent to X or to Y whose score differences are positive. Then the first arrow in L is removed, tested for legality, and if the tests pass, added to the graph; the selected adjacent edges are then reoriented, and arrows for edges adjacent to the endpoints of that edge are added to L if their score differences are positive. This continues until the forward stage finishes, which occurs when the list L is empty. The backward phase is then commenced, populating the list L and proceeding until L is empty. One reports the final graph of this procedure. This implementation scales easily to 1000 variables and, with great patience, to 5000 variables for sparse graphs.

There is still quite a lot of redundant scoring. If one scores Y with no parents, and then scores X->Y, the score for Y with no parents is scored twice. However, eliminating this redundancy for large models requires expensive maps, which we have not pursued successfully here.

Instead, the scoring of the algorithm has been sped up through parallelization. In the above algorithm, there are two steps that lend themselves to parallelization. First, there is the initial step of adding a single edge to the graph. For this, each single edge must be evaluated, and these operations can be done in any order, independently even, with the same result. Since it is a very expensive step, it is a good candidate for parallelization. The other step that one can parallelize to good effect is the scoring of additional edges after the addition of each edge to the model. Once one adds X->Y to the graph, for each W in the graph other than X or Y, one must reevaluate W->X, X->W, W->Y, and Y->W. These are independent checks. With these steps parallelized, the algorithm can be scaled up to 35,000 variables on a 16 GB, 8-processor Macbook Pro with patience (1.6 hours).

Scaling up further is possible, although in doing so, one assumption in particular of GES may usefully be strengthened. GES, as defined, does not assume faithfulness per se but something weaker, more akin to triangle faithfulness (Spirtes and Zhang, 2014). GES can actually discover the model A->B->C->D, A->D, even if the A->B->C->D path exactly cancels the A->D path. This assumption is strengthened to a weak version of faithfulness when doing regressions for scoring. If we know that A has zero total effect on D, we can remove A as a parent when regressing D on its parents (A and C). In that case, we can then do a preprocessing step in which we identify all pairs of variables <X, Y> for which X and Y have zero total effect on other another, and we can simply remove X->Y and Y->X from all scoring regression models. Removing zero correlation edges from the graph allows us to speed up the algorithm considerably. The step of identifying the first edge X->Y to add to the graph is unchanged. Once one adds X->Y, one does not need to consider all W adjacent to X or Y, just those such that <W, X> or <W, Y> are effective connections (i.e., have a non-zero total effect). Doing this on the above-configured MacBook Pro reduces running time by a factor of 5 and saves a good deal of memory. Note that this is primarily a strengthening of assumption for the purpose of scaling up. For smaller models, where more recall is desired, and where one has more time to wait for a result, a version of GES that does not undertake this strengthening is preferable.

The step of rebuilding the pattern after each edge addition can also be optimized. The method outlined above is a global method in which one eliminates all orientations from the graph that do not directly participate in unshielded colliders and then applies the Meek orientation rules to the entire graph. However, reorienting the graph in this way is onerous when the graph becomes large and, moreover, is unnecessary.

One may instead limit revisions of orientation to just the variables directly involved in orientation changes. When one adds X->Y to the graph, one must reorient edges about X and edges about Y and adjust any other orientations in the graph that result. Thus, about X and Y, one can eliminate orientations not directly involved in unshielded colliders and then apply Meek rules to X and Y, and the surrounding variables. Reorienting locally has two implications beyond X and Y and their adjacents. First, new orientations may propagate by the Meek rules applied to adjacent variables; these new orientations must be allowed to propagate and may be quite substantial. Second, if, for instance, triangles are formed in the graph, some orientations adjacent to X and Y may be eliminated, which may result in the elimination of further orientations. Thus, the elimination of orientations must be allowed to propagate as well. Propagating eliminations may be accomplished by identifying edges W->X or W<-W that have been unoriented as W--X by applying the above procedure (reverting to unshielded collider orientations and propagating Meek rules) and then applying the same procedure to W (and similarly for Y). This way, all new orientations and the removal of orientations will be propagated, and the effect on the graph will be the same as if one had reverted the entire graph to unshielded colliders and run the Meek rules. Doing this considerably speeds up the algorithm, by a factor of 5.

One notable point is that the SortedSet collection uses the score difference to order Arrows from largest bump to smallest, but this alone does not account for the possibility that two Arrows may have the same bump. In this case, both Arrows must be added to the list. One may finesse this issue by adjusting the ordering of Arrows in the SortedSet; we order Arrows arbitrarily with the same bump by subtracting the hash codes of the two Arrows. Ordering Arrows with equal scores by hash codes does not produce a complete ordering, but violations are extremely rare, and methods that produce a complete ordering (such as keeping lists of Arrows) are expensive.

6. What does FGS parallelize?

It helps to pause a moment and be clear about what exactly this implementation of the algorithm parallelizes. It only makes sense to run an algorithm on a supercomputer if one has rendered it sufficiently parallel, and GES is not an algorithm that lends itself to easy parallelization. Nevertheless, it can be parallelized fruitfully in a few places, one of which is of great help.

First, the preparation of the covariance matrix can be parallelized by variables. If one precalculates the covariance matrix (which is not the case here), it may be parallelized by striping the matrix by variables in one dimension. We are calculating covariances on the fly, but we use variances heavily, so these deserve precalculation. We may parallelize these by dividing them into groups.

The most effective parallelization is to precalculate the "effect edges" – that is, scores for every edge in the graph in search of the single best edge to add to the graph, as a first step. For sparse graphs, this is the most expensive step in the algorithm, usually taking far and away the most time to complete. It can be parallelized again by ping the lower triangle of edges indexed by the first and second variable in each edge. Parallelization uses almost all the processors for even very large machines; if it does not, the algorithm can be adjusted so that it uses more.

Beyond these, although options for parallelization remain, none is particularly good. The overall structure of the second part of the GES algorithm (which includes most of the algorithmic code) is not especially amenable to parallelization for one simple reason: The overarching idea of GES in the forward phase is to find the best edge to add to the graph, and, when given that first edge, to find the next best edge, and the next best edge, and so on. These are conditional decisions that must be done in serial, not in parallel, and there are a lot of them. However, within each decision, some reasoning may be parallelized to a degree. For instance, in the forward phase, after one has added an edge, one must recalculate quite a number of score differences that may have been affected by the new edge. These are cases where there may be a new parent of a node given the edge or its reverse, in cases where one has not fixed the direction of the edge by implied orientation. These rescorings can be listed, and the list tackled by the rescorer in parallel by simply breaking the list down into smaller lists. This subdividing of the effort may be done both in the forward phase and in the backward phase. The number of processors dedicated to either phase increases from one to four to six through this operation but no further. On our Pittsburgh Supercomputer node with 40 processors, calculating the effect edges easily uses all 40 processors, while the second step uses just four to six processors. Fortunately, the second part of the algorithm (i.e., most of GES) takes much less time to execute than the first, even with both parallelized.

Otherwise, we have not done any more parallelization. Rather, speedups have resulted from overarching strategic choices and detailed profiling work.

7. Simulation Framework.

We aim for FGS to do well on simulation tests of the following form. For each simulation, choose a sparse graph with the number of edges equal to the number of nodes, obtained by adding random forward edges to nodes arranged in a list. One parameterizes each such graph as a structural equation model (SEM), such as this one with the graph X->Y:

$X := eX$
$Y := a X + eY$

In this model, X and Y are continuous variables, eX and eY are disturbances, and a is a linear coefficient. We assume that eX and eY distribute according to some zero-centered Gaussian distribution. The procedure of testing is as follows. First, we generate a random graph by adding random edges in the forward direction, connecting variables in a list. Second, we parameterize this graph as a structural equation model, as above, with coefficients randomly drawn from (-1.5, 0.05) U (0.05, 1.5) and coefficients from U(1, 3), and then draw 1000 samples from the specified distribution recursively and i.i.d. Finally, we run the optimized GES algorithm on this data, comparing edges in the result back to the equivalence class (i.e. the pattern, or CPDAG, see below) of the original graph and reporting back discrepancies in directions and arrow orientations. An upshot of this procedure is that the models we test are linear and Gaussian, with all variables continuous and all data i.i.d. We are testing scale-up under these conditions and have left for future work the problem of testing scale-up when these conditions are not met.

One of the goals of the Center for Causal Discovery is to develop algorithms that can be run on laptop computers as well as on a supercomputer when the capacity of a laptop is insufficient. For laptop testing, we chose a MacBook Pro with two cores (each with two hardware threads, for a total of four processors) and 16G RAM. For high-performance computing testing, we chose a node at the Pittsburgh Supercomputing Center with 40 processors and 384 GB RAM. In both cases, using a machine with more processors would significantly reduce running times and allow us to explore larger models comfortably.

We thought a MacBook Pro would be a representative choice for a laptop machine on which to test the performance of FGS. It is not especially large, even in its maximum configuration. At present, our selected configuration is a middle-of-the-road machine, certainly with no more capacity than one would expect from anyone who might want to use this algorithm. We are careful, as indicated earlier, to calculate covariances on the fly, which is feasible for a thousand samples. Representing covariance matrices directly, even using 4-byte numerical representation, would quickly use up all of the available 16 GB of RAM and would severely limit the size of the problem we could explore. (The maximum number of variables would be about 15,000, even if all 16 GB could be used, which it cannot due to the presence of the operating system.) By calculating covariances on the fly, we need only store the data, and even in 8-byte representation for numbers, this can allow the number of variables to go much higher.

8. Results.

We used run time and precision and recall as the primary measures of performance. We measured both adjacency precision and adjacency recall, as well as arrow (orientation) precision and arrow recall. Two nodes are considered adjacent if they have any type of edge between them. *Adjacency precision* is the number of correctly predicted adjacencies (according to the data-generating model) divided by the number of predicted adjacencies. *Adjacency recall is* the number of correctly predicted adjacencies divided by the total number of adjacencies in the data-generating model. Arrow adjacency is a specialization of adjacency. Two nodes are considered arrow adjacent if they have an arc of the type → between them (in either direction). It is of particular interest, because it represents that two nodes have a direct causal relationship (under assumptions, including causal sufficiency). *Arrow precision* is the number of correctly predicted arrow adjacencies divided by the number of predicted arrow adjacencies. *Arrow recall* is the number of correctly predicted arrow adjacencies divided by the total number of arrow adjacencies in the data generating model.

We tested sparse models of 10,000 to 100,000 variables, with the results shown in Table 1. The precision and recall are above 95% for all tests, which is excellent performance.

Table 1. Results for FGS on the 16 MB, 4 Processor MacBook Pro. "Adj Prec" abbreviates "Adjacency Precision"; "Adj Rec" abbreviates "Adjacency Recall", a similarly for the arrow statistics.

| # Vars | Time (min) | Adj Prec | Adj Rec | Arrow Prec | Arrow Rec |
|---|---|---|---|---|---|
| 10000 | 0.6 | 99.9% | 97.6% | 99.6% | 96.3% |
| 20000 | 2.7 | 99.7% | 97.4% | 99.3% | 96.0% |
| 30000 | 4.4 | 99.6% | 97.3% | 99.0% | 96.0% |
| 40000 | 7.8 | 99.5% | 97.5% | 98.7% | 96.1% |
| 50000 | 12.5 | 99.4% | 97.3% | 98.6% | 95.9% |
| 60000 | 20.5 | 99.3% | 97.4% | 98.3% | 96.1% |
| 70000 | 25.8 | 99.1% | 97.5% | 97.9% | 96.1% |
| 80000 | 32.3 | 99.0% | 97.5% | 97.7% | 96.1% |
| 90000 | 41.1 | 98.9% | 97.6% | 97.5% | 96.2% |
| 100000 | 50.5 | 98.8% | 97.4% | 97.3% | 96.1% |

Compute times increase quadratically with the number of variables. On the MacBook Pro, one eventually reaches the point of using virtual memory. Although the algorithm can continue using higher numbers of variables, we did not explore the upper limit.

By contrast, with the Pittsburgh Supercomputing Center node, it is considerably easier to ascend to larger numbers of variables. In Table 2, we give a table extending the above to 1,000,000 variables. The memory limit for representing covariances directly on this machine was under 200,000 variables; by calculating covariances on the fly, one obtains results for larger numbers of variables. For all results, we use a penalty of 4.

Table 2. Results for FGS on a supercomputer node with 40 processors and 384 MB RAM on a 1,000,000 variable problem, calculating covariances on the fly. See text. "Adj Prec" abbreviates "Adjacency Precision"; "Adj Rec" abbreviates "Adjacency Recall", a similarly for the arrow statistics.

| # Vars | Time (min) | Adj Prec | Adj Rec | Arrow Prec | Arrow Rec |
| --- | --- | --- | --- | --- | --- |
| 10000 | 0.2 | 98.8% | 97.3% | 99.4% | 96.0% |
| 20000 | 0.5 | 99.7% | 97.5% | 99.2% | 96.3% |
| 30000 | 1.0 | 99.6% | 97.5% | 99.0% | 96.2% |
| 40000 | 1.6 | 99.5% | 97.4% | 98.7% | 96.1% |
| 50000 | 2.5 | 99.4% | 97.4% | 98.6% | 96.1% |
| 60000 | 3.1 | 99.2% | 97.4% | 98.1% | 96.0% |
| 70000 | 4.5 | 99.2% | 97.4% | 98.1% | 96.1% |
| 80000 | 6.0 | 99.0% | 97.3% | 97.7% | 95.9% |
| 90000 | 7.1 | 99.0% | 97.5% | 97.5% | 96.2% |
| 100000 | 9.8 | 98.9% | 97.6% | 97.5% | 96.3% |
| 200000 | 34.9 | 97.8% | 97.4% | 94.9% | 96.0% |
| 500000 | 200.0 | 97.5% | 94.7% | 89.0% | 96.1% |
| 1000000[2] | 1070.5 | 99.0% | 93.4% | 99.7% | 97.9% |

For Table 2, precisions and recalls remain high for both adjacencies and arrows, as with Table 1, while time elapsed is understandably much faster across the board.

9. Discussion.

The code for FGS is part of the Tetrad project, www.phil.cmu.edu/tetrad, which has been made open source as of October 28, 2015.[3] The project is at github.com/cmu-phil/tetrad. Code for FGS is in the package edu.cmu.tetrad.search. We do not feel that we have reached the limit of optimization for this problem, but we have gone much further than previous attempts. We would welcome ideas for further optimizing the code. In addition, it is clear that Java is very fast for this problem, despite early dire warnings to the contrary. However, a C implementation would no doubt be faster.

Although not discussed here, the algorithm can be used to analyze discrete variables using a discretely decomposable score, such as BIC or BDeu; we use BDeu, as it is recommended by Chickering (2002). We have scaled this up to 10,000 variables, with excellent precision and fair recall. (As a cautionary note, the accuracy very much depends on the choice of values for hyperparameters of the score.) Work is underway to scale up this discrete score further: for large sample sizes, AD trees will be incorporated, and we are looking into a different technique for increasing the number of variables.

---

[2] Due to time constraints the point for 1,000,000 samples was taken from a run using an earlier version of the code. This will be corrected soon. For the run listed, the penalty discount was increased to 8; the penalty discount for the other runs was 4.

[3] Code for the Tetrad repository was available publically at a location that was difficult to find, as a zip file. Hopefully making the code public in GitHub and converting the project to a Maven repository will help others obtain and work on the code.

Also, although not examined here, FGS handles somewhat denser models relatively easily, though this does slow down the algorithm. Nevertheless, based on experiments, accuracy is good even when the number of edges is two or even three times the number of nodes.

Finally, a fast and accurate pattern search such as FGS can also serve as a component in modifications of other algorithms, such as FCI (Spirtes et al., 2000), with improvements in speed and accuracy. We have successfully performed such combinations and hope to make available modified algorithms with these changes soon.

In short, FGS is a very fast and accurate greedy search for the continuous variable case, suitable for application to large data sets. One side comment is that in applying FGS to large data sets, one may discover the existence of massive hubs. It is reasonable to suspect that such hubs exist in graphs, the structure of which is assumed to be scale-free, such as biological networks. As long as edges point away from these hubs, FGS works fairly well. The existence of such hubs slows down the algorithm to an extent, so the number of variables that we could comfortably analyze might be more on the order of 300,000 or 400,000 rather than 1,000,000, but still within the range of many big data problems.